%% file: root.tex
\documentclass[letterpaper, 10 pt, journal, twoside]{ieeetran}

\usepackage{mathtools}
\usepackage{bbding}

\usepackage{cite}

\usepackage[pdftex]{graphicx}
\graphicspath{{figures/}}
\DeclareGraphicsExtensions{.pdf,.jpeg,.jpg,.png}

\usepackage{amsmath}

\usepackage{url}

\usepackage{subfiles}

\usepackage{xcolor}
\usepackage{booktabs}
\usepackage{tabularx,ragged2e}

\usepackage{hyperref}
\usepackage[nameinlink,noabbrev,capitalise]{cleveref}

\usepackage{comment}

\usepackage[normalem]{ulem}

\definecolor{CommentRed}{rgb}{1.0,0,0}

\title{A Holistic Motion Planning and Control Solution \\to Challenge a Professional Racecar Driver}

\author{Sirish~Srinivasan$^{*}$, Sebastian~Nicolas~Giles$^{*}$, Alexander~Liniger%
\thanks{Manuscript received: February, 24, 2021; Revised June, 11, 2021; Accepted July, 14, 2021. This paper was recommended for publication by Editor Stephen J. Guy upon evaluation of the Associate Editor and Reviewers' comments.}
\thanks{Sirish Srinivasan and Sebastian Nicolas Giles are with AMZ Driverless, ETH Z\"urich, 8092 Z\"urich, Switzerland (e-mail: sirishs@ethz.ch; sgiles@ethz.ch)}%
\thanks{Alexander Liniger is with the Computer Vision Lab, ETH Z\"urich, 8092 Z\"urich, Switzerland (e-mail: alex.liniger@vision.ee.ethz.ch)}
\thanks{$^*$ The authors contributed equally to this work.}
\thanks{A supplementary video providing a high level overview of the approach, along with a demonstration of the experimental results is available.}
\thanks{Digital Object Identifier (DOI): see top of this page.}
}%

\begin{document}

\maketitle
\begin{abstract}
We present a holistically designed three layer control architecture capable of outperforming a professional driver racing the same car. Our approach focuses on the co-design of the motion planning and control layers, extracting the full potential of the connected system. First, a high-level planner computes an optimal trajectory around the track, then in real-time a mid-level nonlinear model predictive controller follows this path using the high-level information as guidance. Finally a high frequency, low-level controller tracks the states predicted by the mid-level controller. Tracking the predicted behavior has two advantages: it reduces the mismatch between the model used in the upper layers and the real car, and allows for a torque vectoring command to be optimized by the higher level motion planners. The tailored design of the low-level controller proved to be crucial for bridging the gap between planning and control, unlocking unseen performance in autonomous racing. The proposed approach was verified on a full size racecar, considerably improving over the state-of-the-art results achieved on the same vehicle. Finally, we also show that the proposed co-design approach outperforms a professional racecar driver.

\end{abstract}

\begin{IEEEkeywords}
Motion and Path Planning, Field Robots, Intelligent Transportation Systems.
\end{IEEEkeywords}

\IEEEpeerreviewmaketitle

\section{Introduction}
\label{sec:intro}
\subfile{sections/introduction.tex}

\section{Low Level Control Design}
\label{sec:llc_formulation}
\subfile{sections/llc_formulation.tex}

\section{Model}
\label{sec:model}
\subfile{sections/vehicle_model.tex}

\section{High Level Control Formulation}
\label{sec:hlc_formulation}
\subfile{sections/hlc_formulation.tex}

\section{Results and Discussion}
\label{sec:results}
\subfile{sections/results.tex}

\section{Conclusion}
\label{sec:conclusion}
\subfile{sections/conclusion.tex}

\section*{Acknowledgment}

We would like to thank the entire AMZ Driverless team, this work would not have been possible without the effort of every single member, and we are glad for having the opportunity to work with such amazing people. We would also like to thank the numerous alumni for the insightful discussions.


\bibliographystyle{IEEEtran}
\bibliography{root.bbl}

\end{document}

%% file: sections/introduction.tex
\IEEEPARstart{A}{utonomous} driving has progressed massively over past few decades, from its humble beginnings in the 1980s \cite{dickmanns1987,Pomerleau1989}, over the DARPA challenges \cite{buehler2005,buehler2009}, to the self-driving car companies of today. One goal for autonomous driving has always been to surpass human driving capabilities. This is especially true for autonomous racing, where professional racecar drivers are a challenging benchmark.
However, most existing methods fall short of this goal \cite{hermansdorfer2020benchmarking}.
In the last years, several motion planning methods for autonomous racing have emerged \cite{Gerdes2012,Liniger2015}. In this paper, we introduce a holistic view-point on motion planning and control of autonomous racecars, and show that the co-design of all layers from track-level trajectory planning to low-level control of the vehicle dynamics results in an unseen performance on a full-sized autonomous racecar. In fact our proposed approach achieves higher driving performance and lower lap times than a professional racecar driver, both driving the same Formula Student Driverless car developed by AMZ Racing, from ETH Z\"urich.

Most autonomous racing motion planners and controllers can be divided into three levels. The first level is track-level planning, where the race line around the track is determined. This can be done using either lap-time optimization methods \cite{lot2014curvilinear,Rucco2015,vazquez2020optimization}, or using the center line \cite{Liniger2015,Kabzan2019_AMZ,Rosolia2017}. The mid-level is tasked to follow the track-level path, and is normally based on two common approaches - static feedback controllers \cite{Gerdes2012,Betz2019,TALVALA2011137} or online optimization-based methods \cite{Liniger2015,funke2016collision,caporale2019towards,williams2016aggressive} such as Nonlinear Model Predictive Control (NMPC). The last level is the Low-Level vehicle Control (LLC) which handles steering and stability control, and typically hosts torque vectoring algorithms which are also beneficial for a human driver \cite{human_llc}.
Even though this layer is fundamental, it is often under-explored, especially the coupling with the mid-level.
There exist several works that make use of a sophisticated LLC. However, these are either designed for human drivers \cite{Kabzan2019_AMZ,vazquez2020optimization} or do not exploit that the level above is an automatic controller \cite{TALVALA2011137,Chatzikomis_2018}. 
In this work, we show that co-designing the LLC to work in harmony with the higher levels allows improving the performance of the autonomous racecar drastically. We achieve this by tracking parts of the mid-level NMPC state trajectory with the LLC. This reinforces recent work which showed that better coupling the track and mid-level controllers \cite{vazquez2020optimization,Novi2019} can improve the performance, and \cite{Chatzikomis_2018} that highlighted the benefits of torque vectoring for autonomous cars. 

\begin{figure}[t]
\begin{center}
\includegraphics[width=0.8\columnwidth] {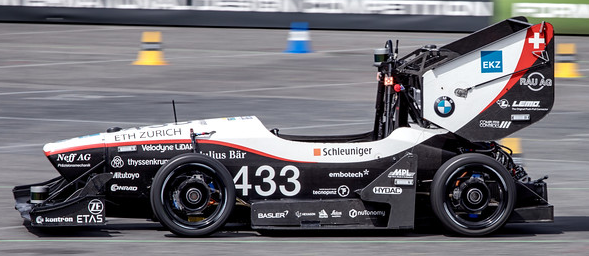}
\end{center}
	\caption{\emph{pilatus} driverless, the formula student race car \copyright FSG - Schulz.}
	\label{fig:pilatus_driving}
\end{figure}

A different view point supporting the coupling of different levels is based on model mismatch. Several papers discovered model mismatch as a crucial issue in autonomous racing - the problem arises due to the relatively simple models used in most autonomous racing stacks. Solutions range from using complex models \cite{Novi2019}, stochastic MPC \cite{carrau2016efficient}, to NMPC with model learning \cite{Kabzan2019_learning,Brunner2017} to learn the model mismatch. All these methods tackle the problem in the mid-level and come with drawbacks in terms of the computational load. However, using our co-design approach, we can use the LLC to make the real-car behave closer to the model of the NMPC.

\begin{figure*}[t]
\centering
\includegraphics[width=0.8\textwidth] {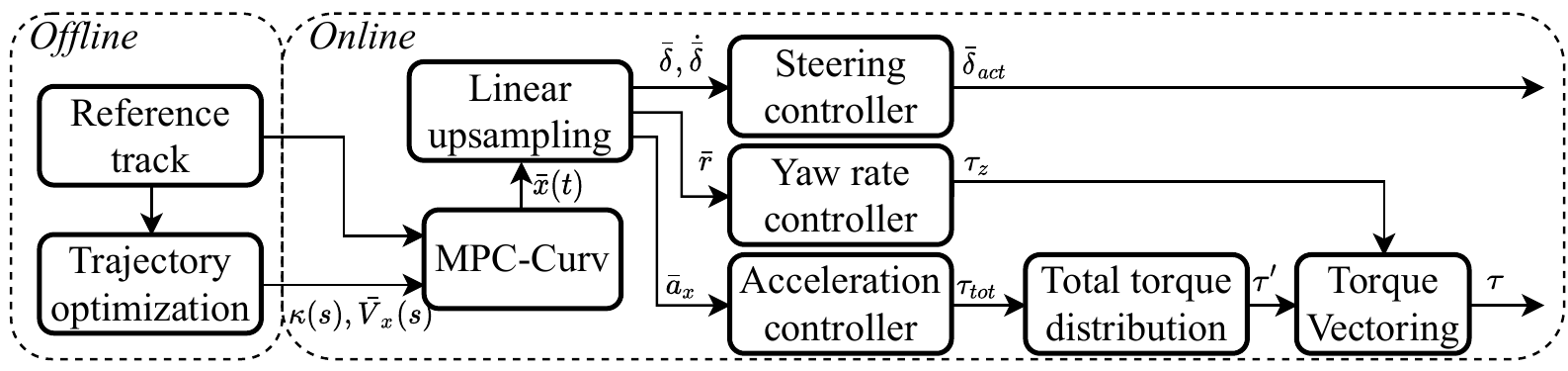}
\caption{Full planning and control architecture.
The feedback from the state estimator is omitted for simplicity. 
}
\label{fig:tv}
\end{figure*}

In this paper we extend the approach proposed in \cite{vazquez2020optimization} and highlight the importance of properly coupled high and low-level controllers. Our contributions are threefold: 
\begin{itemize}
    \item We propose a new low-level controller (LLC) designed to distribute the motor torques of our all-wheel drive race car to reduce the model mismatch between the real car and the model used in the higher level layers.
    \item We propose to directly optimize over a torque vectoring input in the higher level controllers, while interfacing it with the LLC in terms of a yaw-rate target trajectory. This allows to extract the full potential out of our LLC.
    \item We show that the proposed framework can drastically improve the performance over the current state-of-the-art autonomous racing systems, both in simulation and experiments. In fact our method is the first which performs on par with a professional racecar driver, even outperforming the driver in our experimental setup.
\end{itemize}
In this work, attention is given to the LLC formulation and its coupling with the upper control layers, complementing \cite{vazquez2020optimization} which focused on the high and mid-level and their coupling. 
This is also the main difference to other NMPC-based methods \cite{Liniger2015,Rosolia2017,williams2016aggressive} which do not focus on the LLC or torque vectoring. 
With respect to other works that considered LLC, ours is distinguished by the novel interface between the levels:
in \cite{Chatzikomis_2018,roborace_llc,vazquez2020optimization,Kabzan2019_AMZ} the low-level targets are mainly determined by the steering angle, in contrast, we incorporate the yaw rate of the mid-level NMPC. 
In \cite{TALVALA2011137, Gerdes2012}, the focus is on traction control which is a subtask of our LLC.

\section{Co-designed Controller Architecture}

As discussed in the introduction, we build upon \cite{vazquez2020optimization} but introduce several fundamental changes highlighted in our first two contributions. However, we keep a similar architecture for the track and mid-level layers, which are the focus of \cite{vazquez2020optimization}. Our full architecture is shown in \cref{fig:tv}. We assume a reference path from a mapping run and offline compute an optimal trajectory around the track using our Trajectory Optimization (TRO) module. This path is then followed by the MPC-Curv module, using the terminal velocity constraint from the TRO module. However, in contrast to \cite{vazquez2020optimization}, the optimal solution from MPC-Curv is then translated to set point trajectories for the acceleration, yaw rate and steering, which are tracked using our new LLC. 
At the same time, the redeveloped vehicle model makes the motion planning levels aware of the torque vectoring capability of the car and enables optimization over a new yaw moment command. In addition to these two main differences from \cite{vazquez2020optimization}, we also modify the TRO and MPC-Curv modules, resulting in a drastically improved driving performance.

%% file: sections/llc_formulation.tex
\newcommand*\target[1]{\bar{#1}}

Our fully autonomous racecar (see \cref{sec:car_specs} for complete specifications) is equipped with four wheelhub motors that can be controlled independently. Computing the references for each motor creates the need for a LLC, but also allows for significant freedom in its design.
Going beyond basic wheel torque distribution and stabilization\cite{Kabzan2019_AMZ, roborace_llc}, we define a novel input interface for our LLC, consisting of the desired short-term trajectories of key vehicle states, namely the longitudinal acceleration $\bar{a}_x$, yaw rate $\bar r$ and steering angle $\bar \delta$.
Note that a bar on top of a variable denotes a target, while no bar refers to the actual variable.
Importantly, the drive force and torque vectoring commands used by the NMPC and described in Section \ref{sec:curv_model} are purely virtual and are not directly passed to the LLC.
This is the fundamental idea for our high-level to low-level coupling: the state predictions by the NMPC are processed and passed to the LLC which appropriately commands the individual actuators to reproduce them.
The low-level feedback loop actively makes the car behave more like the NMPC model, mitigating the effect of model mismatch in the NMPC.

\subsection{Wheel Torque Controller}

The LLC operates at 200Hz, significantly faster than the 40Hz of the upper level NMPC which sends the target trajectories. To exploit the higher bandwidth, we generate in-between target values using linear up-sampling of the MPC target trajectories. We correct for the delay in the robotic system by slightly shifting the target points in time. 

The LLC computes the torques of the four wheels in a two step approach. First, the required yaw moment and total torque are determined. Second, the individual wheel torques are computed fulfilling these requirements.

The yaw moment is computed using a proportional controller to track the target yaw rate $\tau_{z} = K_P (\target{r}-r)$. The total torque demand is computed using a PID-controller for the target longitudinal acceleration as $\tau_{\text{total}} = PID(\target{a}_x-a_x) + m \target{a}_x + q$, where the feed forward part is designed such that $m$ approximates the inertia of the vehicle including the drive train and $q$ the effect of drag. The computation of $\tau_{z}$ and $\tau_{\text{total}}$ is the main difference to the LLC used in \cite{vazquez2020optimization} and by the human driver in Section \ref{sec:results_exp} where the steering angle is used to compute the target yaw rate \cite{milliken}, and the driver command is directly mapped to the total torque demand.

An initial torque distribution is then computed by splitting the total torque $\tau_{\text{total}}$ equally between the left and right sides of the car. Further, the individual torques are scaled proportional to the normal force $F_z$ on each wheel, resulting in $\tau'=[\tau'_{FL}, \tau'_{FR},\tau'_{RL},\tau'_{RR}]$. The torque vectoring algorithm then determines the torque differences $\Delta\tau_F$ and $\Delta\tau_R$, which adjusts the wheel torques to $\tau=\tau' + [ \Delta\tau_F, -\Delta\tau_F, \Delta\tau_R, -\Delta\tau_R]$. The torque differences $\Delta\tau_F$ and $\Delta\tau_R$ are computed such that the desired yaw moment $\tau_z$ is produced, accounting for the effect of the drive force of the angled front wheels. Since the torque difference neglects the load distribution, in a final step, $\Delta\tau_F$ and $\Delta\tau_R$ are distributed proportional to the vertical tire load on each axle. This allows, for example, to use mainly the rear wheels for torque vectoring during a corner exit. 

The resulting wheel torques are finally tracked by a drive motor controller operating at 1kHz which also implements a traction controller to adapt the reference torque if a slip-ratio-based wheel speed range is violated \cite{ev_traction_control}. 

\subsection{Steering delay compensation module}
The position tracking delay of a servo steering system, due to mechanical compliance and limited power, can severely hinder driving.
To mitigate this we propose a virtual target point $\target{\delta}_{\text{act}}$ as a reference for the steering positioning actuator, based on an external steering shaft sensor measurement $\delta$.
The virtual target point is estimated as 
$\target{\delta}_{\text{act}} = \target{\delta} + K_P (\target{\delta}-\delta) + K_D (\dot{\target{\delta}} - \dot{\delta})$, where $\dot{\target{\delta}}(t)$ is the steering rate from the MPC predictions.

%% file: sections/vehicle_model.tex
Given the LLC, we introduce the vehicle model used in the higher level motion planners. Similar to \cite{vazquez2020optimization} we formulate a dynamic bicycle model in curvilinear coordinates, but we modify the interface with our LLC (See \cref{sec:llc_formulation}). 

\subsection{Curvilinear Dynamic Bicycle Model}
\label{sec:curv_model}
\begin{figure}[h]
    \centering
    \includegraphics[width=0.35\textwidth]{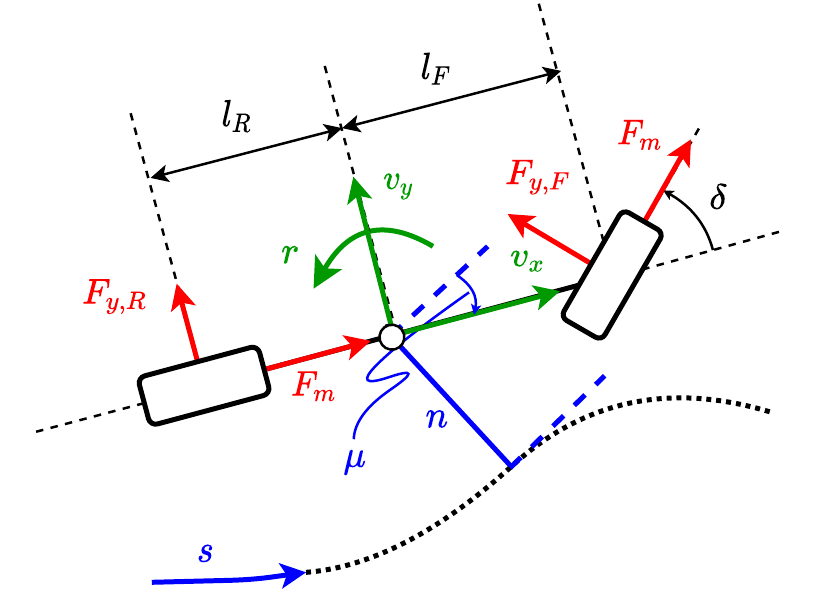}
    \caption{Visualization of the curvilinear coordinates (blue), as well as the dynamic states (green) and the forces (red).}
    \label{fig:model}
\end{figure}

Curvilinear coordinates describe a coordinate frame (Frenet frame) formulated locally with respect to a reference path, drastically simplifies path following formulations. In our case the reference path can be the center line or the track-level optimized path. The kinematic states in the curvilinear setting describe the state relative to the reference path and are the progress along the path $s$, the deviation orthogonal to the path $n$, and the local heading $\mu$. Note that the dynamic states are not influenced by the change in the coordinate system, and in our model we consider the longitudinal $v_x$ and lateral velocities $v_y$, and yaw rate $r$. A visualization of the curvilinear coordinates as well as the other states is shown in \cref{fig:model}. 

Since the LLC handles traction control and considering load changes, we can use a relatively simple model in our higher levels. We follow the popular modeling approach successfully used in \cite{Liniger2015,Kabzan2019_AMZ,vazquez2020optimization}, and uses a dynamic bicycle model with Pacejka tire models. Compared to \cite{vazquez2020optimization} we include some important differences - first we assume that the force of the motors $F_M$ can be directly controlled, and for simplicity assume that the same motor force is applied at the front and rear wheels. This approach is better aligned with our LLC, which is designed to track an acceleration target and does not expect a driver command as in \cite{vazquez2020optimization}. Second, we introduce the torque vectoring moment $M_{tv}$ as an input. This is in contrast to \cite{Kabzan2019_AMZ,vazquez2020optimization} where the torque vectoring was determined by a simple P-controller. This input is fundamental, as it allows the higher level controllers to fully utilize the torque vectoring capabilities, going beyond a simple rule-based method designed for human drivers. We use a state lifting technique to consider input rates, but do not lift $M_{tv}$ to allow the high level controllers to use the torque vectoring for highly transient situations. The state is given by $\mathbf{\tilde{x}} = [s, n, \mu, v_x, v_y, r, F_M, \delta]^T$ and the input as $\mathbf{u} = [\Delta F_M,\Delta \delta,  M_{tv}]^T$. Resulting in the following system dynamics,
\begin{align}
\label{eq:dynamics}
    &\dot{s} = \dfrac{v_x\cos{\mu}-v_y\sin{\mu}}{1-n \kappa(s)}\,, \nonumber\\
    &\dot{n} = v_x\sin{\mu}+v_y\cos{\mu}\,, \nonumber\\
    &\dot{\mu} = r - \kappa(s) \dot{s}\,, \\
    &\dot{v}_x = \tfrac{1}{m}(F_M (1+\cos{\delta}) - F_{y,F} \sin{\delta} + m v_y r - F_{\text{fric}})\,, \nonumber\\
    &\dot{v}_y = \tfrac{1}{m}(F_{y,R} + F_M  \sin{\delta} + F_{y,F} \cos{\delta} - m v_x r)\,,\nonumber\\
    &\dot{r} = \tfrac{1}{I_z}((F_M  \sin{\delta} + F_{y,F} \cos{\delta})l_F - F_{y,R}l_R + M_{tv})\,,\nonumber \\
    &\dot{F}_M = \Delta F_M\,, \nonumber \\
    &\dot{\delta} = \Delta \delta\,, \nonumber
\end{align}
where $l_F$ and $l_R$ are the distances from the Center of Gravity (CoG) to the front and rear wheels respectively, $m$ is the mass of the vehicle and $I_z$ the moment of inertia. Finally, $\kappa(s)$ is the curvature of the reference path at the progress $s$. We denote the dynamics in \eqref{eq:dynamics} as $\mathbf{\dot{\tilde{x}}} = f^c_t(\mathbf{\tilde{x}}, \mathbf{u})$, where the superscript $c$ highlights that it is a continuous model and the subscript $t$ that it is a time-domain model.

The lateral forces at the front $F_{y,F}$ and rear $F_{y,R}$ tires are modeled using a simplified Pacejka tire model \cite{pacejka1992magic},
\begin{align}
\label{eq:lateral_forces}
\begin{split}
    F_{y,F} & = F_{N,F} D\sin{(C\arctan{(B\alpha_F)})}\,, \\
    F_{y,R} & = F_{N,R} D\sin{(C\arctan{(B\alpha_R)})}\,, \\
\end{split}
\end{align}
where $\alpha_F = \arctan{( \frac{v_y + l_F r}{v_x} )} - \delta$ and $\alpha_R = \arctan{( \frac{v_y - l_R r}{v_x})}$ are the slip angles at the front and rear wheels respectively, and $B$, $C$ and $D$ are the parameters of the simplified Pacejka tire model. The net normal load $F_{N,net} = m g + C_l v_x^2$, where $C_l$ is a lumped lift coefficient. Compared to \cite{vazquez2020optimization} we also consider the aerodynamic downforce, which is important since we push the car to the limit of friction. The resulting normal loads on the front and rear tires are given by $F_{N,F} = F_{N,net} l_R /(l_F + l_R) $ and $F_{N,R} = F_{N,net} l_F /(l_F + l_R)$. Finally, the friction force $F_{\text{fric}}$ is a combination of a static rolling resistance $C_r$ and the aerodynamic drag term $C_d v_x^2$.

\subsection{Constraints}
\label{sec:constraints}
Similar to \cite{vazquez2020optimization} we impose constraints to ensure that the car remains within the track, and that we do not demand inputs that violate friction ellipse or input constraints. More precisely we have a track constraint $\mathbf{\tilde{x}} \in \mathcal{X}_{\text{Track}}$, which is a heading-dependent constraint on the lateral deviation $n$, and ensures that the whole car remains inside the track \cite{liniger2020safe}, 
\begin{align}
\label{eq:boundary}
\begin{split}
    n + L_c\sin{|\mu|} + W_c\cos{\mu} & \leq \mathcal{N}_L(s) \,,\\
    -n + L_c\sin{|\mu|} + W_c\cos{\mu} & \leq \mathcal{N}_R(s)\,,
\end{split}
\end{align}
where $L_{c}$ and $W_c$ are the distances from the CoG to the furthest apart corner point of the car, and $\mathcal{N}_{R/L}(s)$ are the left and right track width at the progress $s$. The tire models used \eqref{eq:lateral_forces} do not consider combined slip. To prevent the high level layers to demand unrealistic accelerations from the LLC, we limit the combined forces to remain within a friction ellipse, 
\begin{align}
\label{eq:ellipse}
\begin{split}
(\rho_{long}F_M)^2 + F_{y,F/R}^2 & \leq (\lambda D_{F/R})^2\,, \\
\end{split}
\end{align}
where $\rho_{long}$ defines the shape of the ellipse, and $\lambda$ determines the maximum combined force. We denote the friction ellipse constraints in \eqref{eq:ellipse} by $\mathbf{\tilde{x}} \in \mathcal{X}_\text{FE}$.

Finally, we consider box constraints for both the physical inputs and their rates. We introduce a compact notation for all these inputs $\mathbf{a} = [F_M, \delta, \Delta F_M, \Delta \delta, M_{tv}]^T$, and constrain them to their physical limits by the box constraint $\mathbf{a} \in \mathcal{A}$.

%% file: sections/hlc_formulation.tex
\subsection{Trajectory Optimization (TRO)}
\label{sec:traj_opt}

Given the vehicle model and the constraints, we now focus on the higher level controllers. First the offline track-level trajectory optimization is described. Following \cite{vazquez2020optimization}, we transform the continuous time dynamics \eqref{eq:dynamics} into the spatial domain, with progress $s$ as the running variable instead of time $t$. This transformation can be achieved as follows, 
\begin{align}
\label{eq:spatialdomain}
\begin{split}
&\mathbf{\dot{\tilde{x}}} = \frac{\partial \mathbf{\tilde{x}}}{\partial t} = \frac{\partial \mathbf{\tilde{x}}}{\partial s}\frac{\partial s}{\partial t}\,, \\
& \Rightarrow \frac{\partial \mathbf{\tilde{x}}}{\partial s} = \frac{1}{\dot{s}}f^c_t(\mathbf{\tilde{x}}(s),\mathbf{u}(s))  = f^c_s(\mathbf{\tilde{x}}(s),\mathbf{u}(s)) \,,
\end{split}
\end{align}
where $f^c_s(\mathbf{\tilde{x}}(s),\mathbf{u}(s))$ is the continuous space model. This transformation also makes the $s$ state redundant allowing us to reduce the state to $\mathbf{x} = [n,\mu,v_x,v_y,r,F_M,\delta]^T$.

To formulate the track-level optimization problem, we use the center line reference path and discretize the continuous space model $f^c_s(\mathbf{{x}}(s),\mathbf{u}(s))$. An Euler forward integrator, with discretization $\Delta s$ is used, resulting in the discrete space system $\mathbf{{x}}_{k+1} = f_s^d(\mathbf{{x}}_k, \mathbf{u}_k) = \mathbf{{x}}_k + \Delta s f_s^c({{x}}_k,\mathbf{u}_k)$.
For racing, it is necessary to optimize for a periodic trajectory. Thus, we add a periodicity constraint $f_s^d(\mathbf{{x}}_N, \mathbf{u}_N) = \mathbf{{x}}_{0} $, where $N$ is the number of discretization steps.

The cost function seeks to maximize the progress rate $\dot{s}$, and also contains two regularization terms - a slip angle cost and a penalty on the input rates. The slip angle cost penalizes the difference between the kinematic and dynamic side slip angles as $B(\mathbf{{x}}_k) = q_\beta(\beta_{\rm{dyn},k} - \beta_{\rm{kin},k})^2$,
where $q_\beta > 0$ is a weight, $\beta_{\rm{kin},k} = \arctan (\delta_k l_R /(l_F + l_R))$, and $\beta_{\rm{dyn},k} = \arctan (v_{y,k} / v_{x,k})$. The regularizer on the input rates is $\mathbf{u}^T R \mathbf{u}$, where $R$ is a diagonal weight matrix with positive weights. In summary, the overall cost function is, 
\begin{equation}
\label{eq:tro_cost}
    J_{\rm{TRO}}(\mathbf{{x}}_k, \mathbf{u}_k) = -\dot{s}_k + \mathbf{u}^T R \mathbf{u} + B(\mathbf{{x}}_k) \,.
\end{equation}
Note that we introduced a new cost function compared to \cite{vazquez2020optimization}, which minimized the time. Our new cost function is nearly identical to the one used in the MPC-Curv motion planner (\cref{sec:mpc}), which makes tuning easier, and better aligns the solutions of the two levels.  

Finally, we combine the cost, model and model related constraints from \cref{sec:constraints}, to formulate the trajectory optimization problem,
\begin{equation}
\begin{aligned}
& \underset{\mathbf{{X}},\mathbf{U}}{\text{min}}
    && \sum_{k=0}^{N} J_{\rm{TRO}}(\mathbf{{x}}_k, \mathbf{u}_k) \\
& \; \text{s.t.}
    && \mathbf{{x}}_{k+1} = f_s^d(\mathbf{{x}}_k, \mathbf{u}_k)\,,\\
&&& f_s^d(\mathbf{{x}}_N, \mathbf{u}_N) = \mathbf{{x}}_{0} \,,\\
&&& \mathbf{{x}}_k \in \mathcal{X}_{\text{Track}}, \quad \mathbf{{x}}_k \in \mathcal{X}_{\text{FE}} \,,\\
&&& \mathbf{a}_k \in \mathcal{A}, \quad k = 0,...,N \,,
\label{eq:TRO}
\end{aligned}
\end{equation}
where $\mathbf{{X}} = [\mathbf{{x}}_0,...,\mathbf{{x}}_N]$ and $\mathbf{{U}} = [\mathbf{{u}}_0,...,\mathbf{{u}}_N]$. The problem is formulated using \emph{CppAD} \cite{CppAD} and solved using \emph{Ipopt} \cite{ipopt}.

\subsection{MPC-Curv}
\label{sec:mpc}
We use MPC-Curv to follow the trajectory from TRO. Since the objectives of TRO and MPC-Curv are similar, we reuse large parts of the formulation, including the progress rate maximization. However, since the MPC-Curv problem is solved online, a few changes are necessary - first, we use the time-domain model since it is better suited for online control (see \cite{vazquez2020optimization} for a discussion). Second, since MPC-Curv has a limited prediction horizon, we use the TRO solution for long term guidance. This is done in two ways - first a regularization cost on the lateral deviation $n$ from the TRO path and more importantly a terminal velocity constraint to notify MPC-Curv about upcoming braking spots beyond the prediction horizon. 

To formulate the MPC-Curv problem, we first reduce the state to $\mathbf{x} = [n,\mu,v_x,v_y,r,F_M,\delta]^T$. We use the initial guess of $s$ to evaluate all quantities depending on $s$ such as the curvature $\kappa(s)$ outside the MPC, and hence the $s$-state decouples. Second, we discretize the model using a second order Runge Kutta method, resulting in $f_t^d(\mathbf{x}_t, \mathbf{u}_t)$. In order to decouple the prediction horizon in terms of steps and time, we discretize the system with a sampling time different from the update frequency of the controller. We introduce this as a time scaling factor $\sigma$ which multiplies the sampling time of our controller. Note that this requires interpolating the previous optimal solution to get an initial guess, but at the same time allows to predict further in time with the same horizon in steps. This allows us to run MPC-Curv with higher update rates.

The MPC-Curv cost function is identical to the TRO \eqref{eq:tro_cost}, with the addition of a path following cost on $n$, 
\begin{equation}
J_{\rm{MPC}}(\mathbf{{x}_t}, \mathbf{u}_t) = -\dot{s}_t + q_n n_t^2 + \mathbf{u}^T R \mathbf{u} + B(\mathbf{x}_t) \,,\\
\end{equation}
where $q_n$ is a positive regularization weight. Thus, we can now formulate the MPC-Curv problem, 

\begin{equation}
\label{eq:MPC}
\begin{aligned}
& \underset{\mathbf{X}, \mathbf{U}}{\text{min}}
    && \sum_{t=0}^{T} J_{\rm{MPC}}(\mathbf{x}_t, \mathbf{u}_t) \\
& \; \text{s.t.}
    && \mathbf{x}_0 = \mathbf{\hat{x}}\,,\\
    &&& \mathbf{x}_{t+1} = f_t^d(\mathbf{x}_t, \mathbf{u}_t) \,,\\
&&& \mathbf{{x}}_t \in \mathcal{X}_{\text{Track}}, \quad \mathbf{{x}}_t \in \mathcal{X}_{\text{FE}} \,,\\
&&& \mathbf{a}_t \in \mathcal{A}, \quad v_{x,T} \leq \bar{V}_{x}(s_T)\,, \\
&&& t = 0,...,T\,,
\end{aligned}
\end{equation}
where the subscript $t$ is used to highlight that the problem is formulated in the time domain, and $T$ is the prediction horizon. Further, $\mathbf{\hat{x}}$ is the current curvilinear state estimate and $v_{x,T} \leq \bar{V}_{x}(s_T)$ the terminal constraint from the TRO solution. The optimization problem is solved using ForcesPro \cite{FORCESPro,FORCESNLP}.

%% file: sections/results.tex
We first discuss our robotic platform - the autonomous racecar \emph{pilatus driverless}, followed by implementation details. Thereafter, we benchmark our control approach in simulation using a realistic vehicle simulator to study the low-level adaptations. Finally, we present experimental results including an in-depth comparison with a professional racing driver. 

\subsection{pilatus driverless}
\label{sec:car_specs}
All our experiments are performed using the autonomous racecar \emph{pilatus driverless} (shown in \cref{fig:pilatus_driving}). \emph{pilatus} is a lightweight single seater race car, with an all-wheel drive electric powertrain. The racecar can produce $\pm375$\,Nm of torque at each wheel by means of four independent $38.4$\,kW motors. Our racecar can accelerate from $0-100$\,km/h in $2.1$\,s and can reach lateral accelerations of over 20\,m/s$^2$. \emph{pilatus} is equipped with a complete sensor suite including two LiDARs, three cameras, an optical ground-velocity sensor and two IMUs. The low-level control as well as the state estimation are deployed on an ETAS ES900 real-time embedded system; the remainder of the Autonomous System (AS), including mapping and localization, runs on an Intel Xeon E3 processor, see \cite{valls2018design,gosala2019redundant,Kabzan2019_AMZ,srinivasan2020end,andresen2020accurate} for more details. 

\subsection{TRO and MPC-Curv Implementation Details} \label{sec:TRO_imp}
TRO uses a spatial discretization of $\Delta s = 0.5$\,m. Given this discretization, the optimization problem defined in \eqref{eq:TRO} is solved in $\sim 5$\,s on an Intel Xeon E3 processor. We run MPC-Curv at a frequency of 40Hz, with a time scaling $\sigma = 1.5$. We use a prediction horizon of $T=40$ which results in a time horizon of 1.5\,s, using the time scaling. 

\subsection{Simulation Study}
\label{sec:results_sim}

\begin{figure}[h]
\begin{center}
\includegraphics[width=0.8\columnwidth] {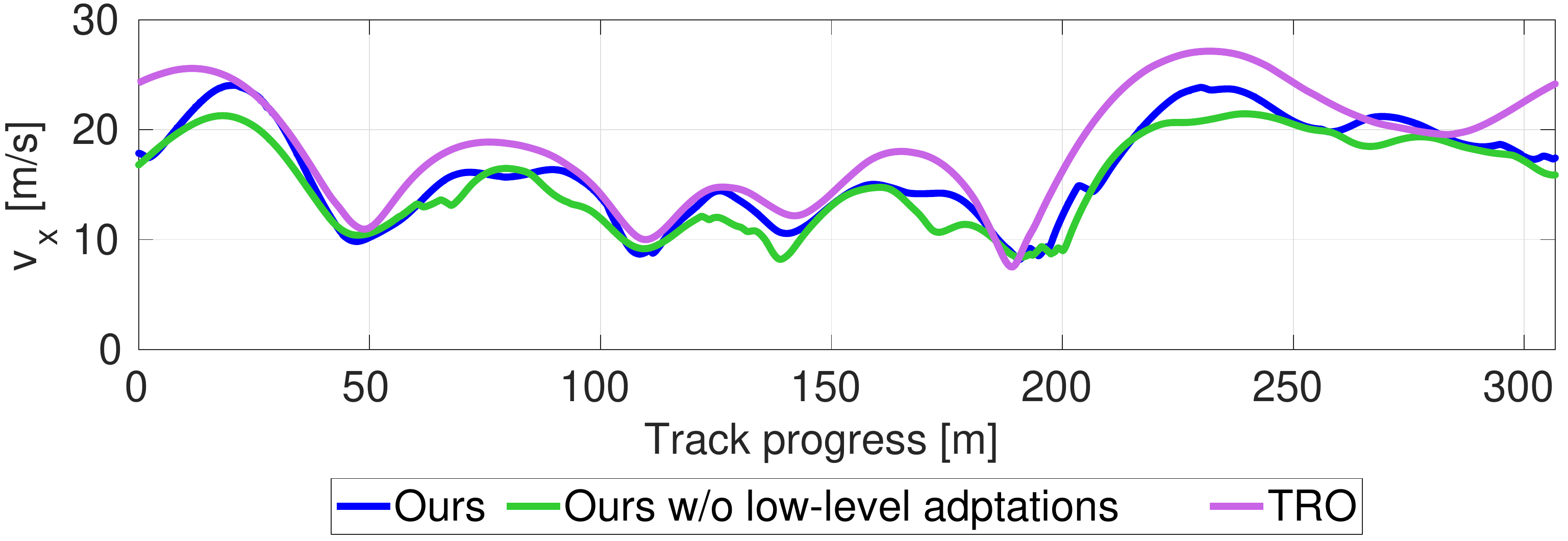}
\end{center}
	\caption{Simulated longitudinal velocity against track position, comparing our approach with and w/o the proposed low-level adaptations, against TRO.}
	\label{fig:sim_vx_progress}
\end{figure}

To highlight the benefits achieved by co-designing the high and low-level controllers, we compare our full pipeline against a modified version of it that cannot optimize over torque vectoring and uses the more basic LLC as in \cite{vazquez2020optimization}. Note that the second method is similar to \cite{vazquez2020optimization}. We perform a simulation on the Formula Student Germany racetrack from 2018, using a high fidelity vehicle dynamics simulator. As an upper performance bound we also include the TRO solution, which uses the dynamic bicycle model \eqref{eq:dynamics} with no mismatch. \cref{fig:sim_vx_progress} shows the longitudinal velocity $v_x$ against the progress along the track. From the comparison, it is clearly visible that our approach can reach higher speeds. The lap times of the three methods confirm this: our full system completes one lap in 19.9\,s whilst without the low-level adaptations the lap time is 22.1\,s, for reference, the TRO optimal lap time is 18.0\,s.

\subsection{Experimental Comparison}
\label{sec:results_exp}

For our comparison with a professional racing driver, we setup a race track compliant with the Formula Student Driverless regulations \cite{fsg-rules}, composed of sharp turns, straights and chicanes.
We tried to keep the comparison between the human driver and our proposed solution as fair as possible, however, there are some differences. First, in autonomous mode, the car is lighter due to the absence of a driver. Second, for safety, the AS speed was limited to 18\,m/s, whereas the human driver was unrestricted and reaches speeds up to 22\,m/s. Third, the AS can only use regenerative braking, since the hydraulic brakes are exclusively used for an emergency braking system. Since this is not necessary for the human driver, the hydraulic brakes can also be actuated, allowing higher deceleration. We would also like to note that the steering system used by the AS is slower than the steering actuated by a driver. 

The experiments were run consecutively, with one run for each of the driving modes. The duration was of $12$ and $18$ laps for the human and AS respectively. In both cases, the car remained within the track and did not hit any cones.

\subsubsection{Lap-time comparison}
\label{sec:laptime_comp}
\begin{figure}[h]
\centering
\includegraphics[width=0.95\columnwidth] {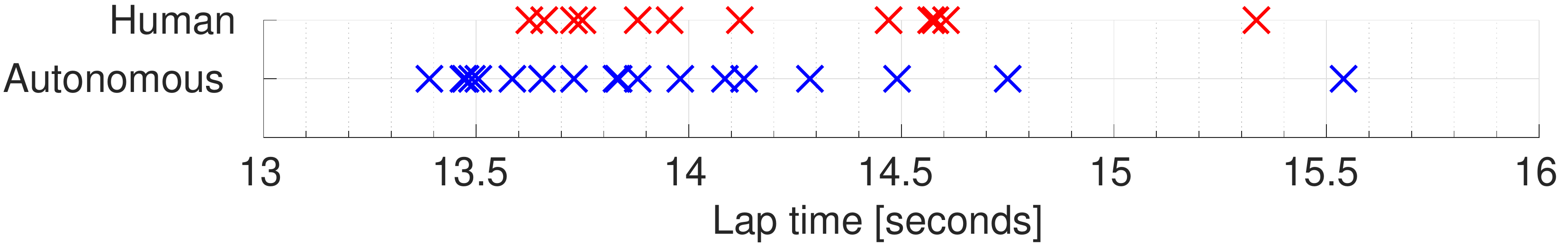}
	\caption{Lap-time distribution for autonomous system and human driver.}
	\label{fig:laptime_comp}
\end{figure}

All of the lap times recorded during the experiment are shown in \cref{tab:laptime_comp} and \cref{fig:laptime_comp}, no data was discarded.
Our proposed controller achieves both the lowest average and minimum lap times. In \cref{fig:laptime_comp} we can see that the autonomous system achieves six laps that are faster than the fastest lap by the professional human driver. Note that the variability of lap times in the autonomous mode comes from a few instances where an emergency planner is triggered and automatically slows down the car to avoid leaving the track. This can occur during large combined slip (corner exit), an issue we want to tackle in future work.

\begin{table}[h]
\centering
\caption{Lap-time comparison between human and autonomous driver}
\label{tab:laptime_comp}
\renewcommand\baselinestretch{1.0}
\begin{tabularx}{0.32\textwidth}{l c c}
    \toprule
    & Human & Autonomous \\
    \midrule
    Best lap-time(s) & 13.62 & 13.39 \\
    Mean lap-time(s) & 14.19 & 13.95 \\
    \bottomrule
\end{tabularx}
\end{table}

\subsubsection{Driving comparison}
\label{sec:drive_comp}

\begin{figure}[h]
\centering
\includegraphics[width=0.8\columnwidth] {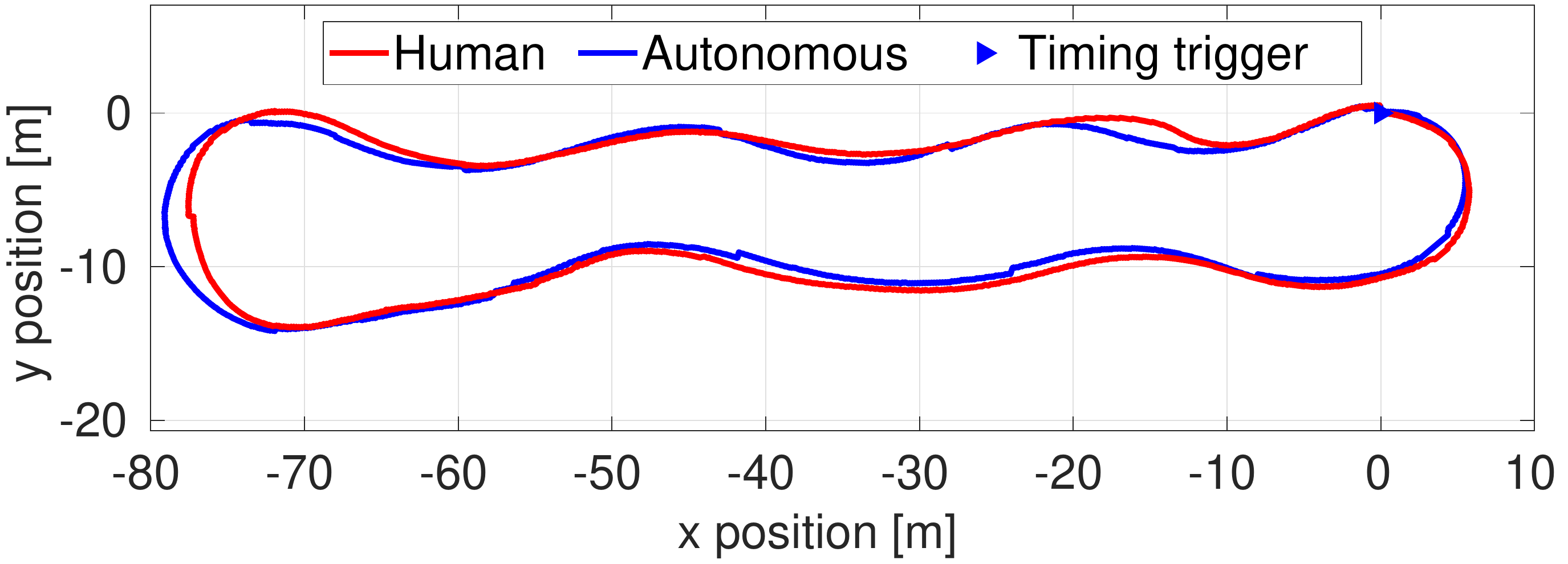}
	\caption{GPS trajectory comparison of the best autonomous and human driven laps. The driving direction is clockwise.}
	\label{fig:path_comp}
\end{figure}

\begin{figure}[h]
\centering
\includegraphics[width=0.8\columnwidth] {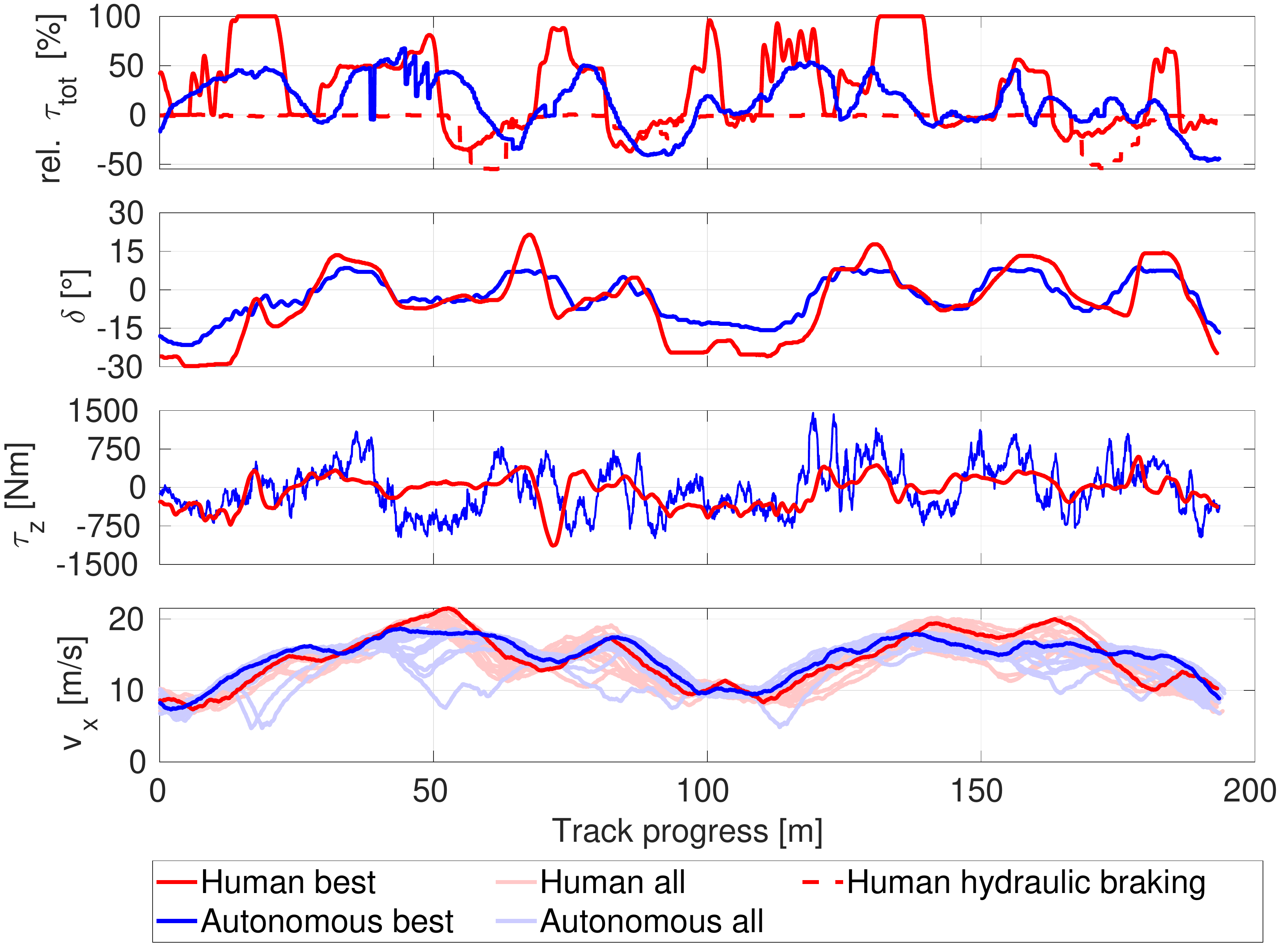}
	\caption{Comparison of the vehicle dynamics on the best lap. The velocity in all of the laps is also shown in a lighter color.}
	\label{fig:lap_comp}
\end{figure}

\cref{fig:path_comp} shows the comparison of the driven GPS paths for the autonomous and human modes, and \cref{fig:lap_comp} compares the longitudinal velocity of all driven laps. One major difference between the driven paths can be seen at the increasing radius curve on the left extreme. The AS does not follow the intuitive inner radius of the curve but goes wide, which allows later braking and a faster and straighter curve exit, which we can see in \cref{fig:lap_comp} at 100\,m. A similar difference can also be seen in the curve at the right extreme, where especially the last chicane before the curve entry and the curve exit are driven faster. The offline TRO problem can efficiently perform such trade-offs between travelled distance and speed, while even a professional driver needs significant track time to evaluate such trade-offs. 

The other significant difference we can see in \cref{fig:lap_comp} is the effect of the 18\,m/s speed limiter for the AS. However, our controller is able to brake later and accelerate earlier at several locations along the lap, which in total results in lower lap-times, even with the top speed disadvantage. 

\subsubsection{Inputs comparison}
\label{sec:veh_comp}
The low-level inputs are shown in \cref{fig:lap_comp}. The AS in contrast to the human driver, makes limited use of the steering. 
This is compensated using more torque vectoring, which provides a faster response of the lateral dynamics. This would not be possible without the strong coupling at the core of our holistic architecture. Further, the high-frequency component of the yaw moment is the intended effect of the LLC tracking to compensate for model mismatch.

The final difference is the relative total torque input, where we can see that the AS demands less torque. This is due to the speed limit, and the fact that the friction ellipse constraint limits the torque before the traction controller. The human driver on the other hand relies on the traction controller in certain situations, e.g., at 100\,m. 

\subsubsection{Maximum performance}
\label{sec:gg}

\begin{figure}[h]
\begin{center}
\includegraphics[width=0.9\columnwidth] {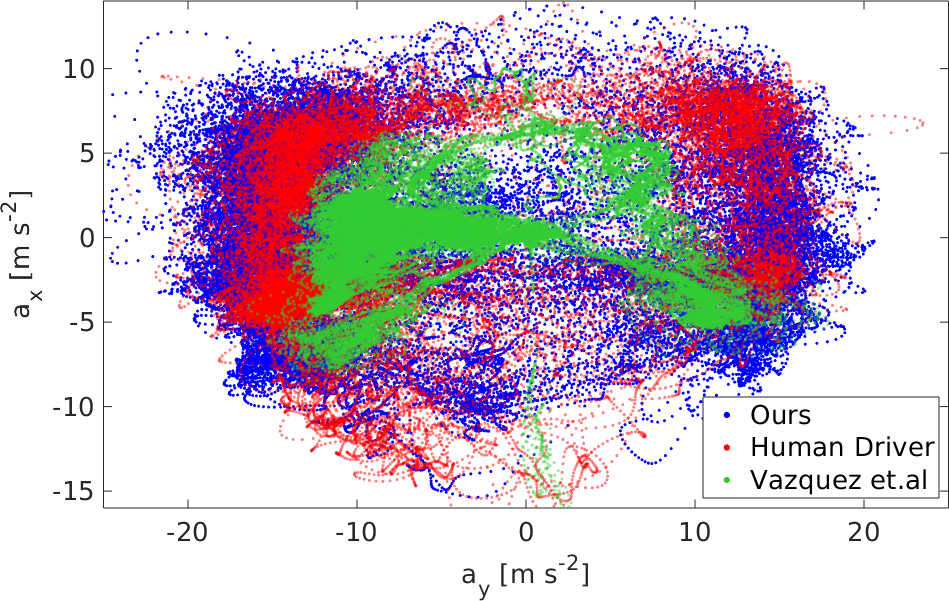}
\end{center}
	\caption{Longitudinal and lateral acceleration comparison between our system, the professional human driver and Vazquez et.al \cite{vazquez2020optimization}}
	\label{fig:gg}
\end{figure}

\cref{fig:gg} compares the longitudinal and lateral accelerations recorded by the car in autonomous and human driven modes. We can see that the highest lateral, positive longitudinal and combined accelerations are achieved by the AS. The human driver has a higher negative longitudinal acceleration, due to the availability of hydraulic brakes, which are not used by the AS. Finally, we can also see the drastic performance difference between our approach and \cite{vazquez2020optimization}, while using the same car.

%% file: sections/conclusion.tex
In this paper we proposed a holistic way to think about motion planner and controller design for autonomous racing.
The idea is that all hierarchical control layers should be designed while keeping the other layers in mind.
We proposed a low-level controller that actuates the steering, and distributes the wheel torques to track the acceleration, yaw rate and steering trajectories predicted by a higher level NMPC.
The higher level motion planners consider the torque vectoring capabilities of the low-level controller. Thus, the model mismatch between the levels can be reduced while the capabilities of the car can be fully extracted. 
We show this by comparing the performance of our autonomous controller with a professional human driver both driving the same full-sized autonomous racecar. 
Our autonomous controller is able to better the driver both in peak and average lap-times. Future work will include real-time identification of the peak tire performance to benefit from the full grip potential, and data-driven learning of the cost function.